\documentclass{article}

\usepackage{microtype}
\usepackage{graphicx}
\usepackage{subfigure}
\usepackage{colortbl}%
\usepackage{appendix}

  \renewcommand\thesection{\arabic{section}}
\usepackage{booktabs}

\usepackage{hyperref}

\usepackage[dvipsnames]{xcolor}

\newcommand{\ourmethod}{GALA3D}

\newcommand*\case{\textcolor{green}{%
  \ensuremath{\blacksquare}}}

\newcommand*\casee{\textcolor{red}{
  \ensuremath{\blacksquare}}}

\newcommand*\caseee{\textcolor{blue}{
  \ensuremath{\blacksquare}}}

\newcommand*\caseeee{\textcolor{lime}{
  \ensuremath{\blacksquare}}}

\newif\ifdrafting
\draftingtrue
\draftingfalse
\ifdrafting

\else
    
\fi

\definecolor{tabfirst}{rgb}{1, 0.7, 0.7} 
\definecolor{tabsecond}{rgb}{1, 0.85, 0.7} 
\definecolor{tabthird}{rgb}{1, 1, 0.7} 

\newcommand{\aftertab}{\vspace{-1em}}

\newcommand{\aroundeqn}{\vspace{-.2em}}

\usepackage[accepted]{icml2024}

\usepackage{amsmath}
\usepackage{amssymb}
\usepackage{mathtools}
\usepackage{amsthm}
\usepackage{bbm}
\usepackage{newtxmath}
\usepackage{tikz}
\usepackage{gensymb}
\usepackage{graphicx}
\usepackage{lineno}
\usepackage{mathtools}
\usepackage{microtype}
\usepackage{multirow,bigdelim}
\usepackage{soul}
\usepackage{subfigure}
\usepackage{xspace}
\usepackage{array,makecell}
\usepackage{booktabs}
\usepackage{caption}
\usepackage{float}
\usepackage{hyperref}
\usepackage[capitalize,noabbrev]{cleveref}
\usepackage{paralist}

\usepackage[capitalize,noabbrev]{cleveref}

\theoremstyle{plain}

\theoremstyle{definition}

\theoremstyle{remark}

\begin{document}

\twocolumn[{

\icmltitle{\ourmethod{}: Towards Text-to-3D Complex Scene Generation via Layout-guided Generative Gaussian Splatting}

\begin{icmlauthorlist}
\icmlauthor{Xiaoyu Zhou}{yyy}
\icmlauthor{Xingjian Ran}{yyy}
\icmlauthor{Yajiao Xiong}{yyy}
\icmlauthor{Jinlin He}{yyy}
\icmlauthor{Zhiwei Lin}{yyy} \\
\icmlauthor{Yongtao Wang}{yyy,ttt}
\icmlauthor{Deqing Sun}{comp}
\icmlauthor{Ming-Hsuan Yang}{comp,sch}
\end{icmlauthorlist}

\icmlaffiliation{yyy}{Wangxuan Institute of Computer Technology, Peking University}
\icmlaffiliation{comp}{Google DeepMind}
\icmlaffiliation{sch}{University of California, Merced}
\icmlaffiliation{ttt}{National Key Laboratory for Multimedia Information Processing}

\icmlcorrespondingauthor{Yongtao Wang}{wyt@pku.edu.cn}

\icmlkeywords{Machine Learning, ICML}

\vskip 0.3in

\begin{center}
  \vspace{-5mm}
  \includegraphics[width=0.99\linewidth]{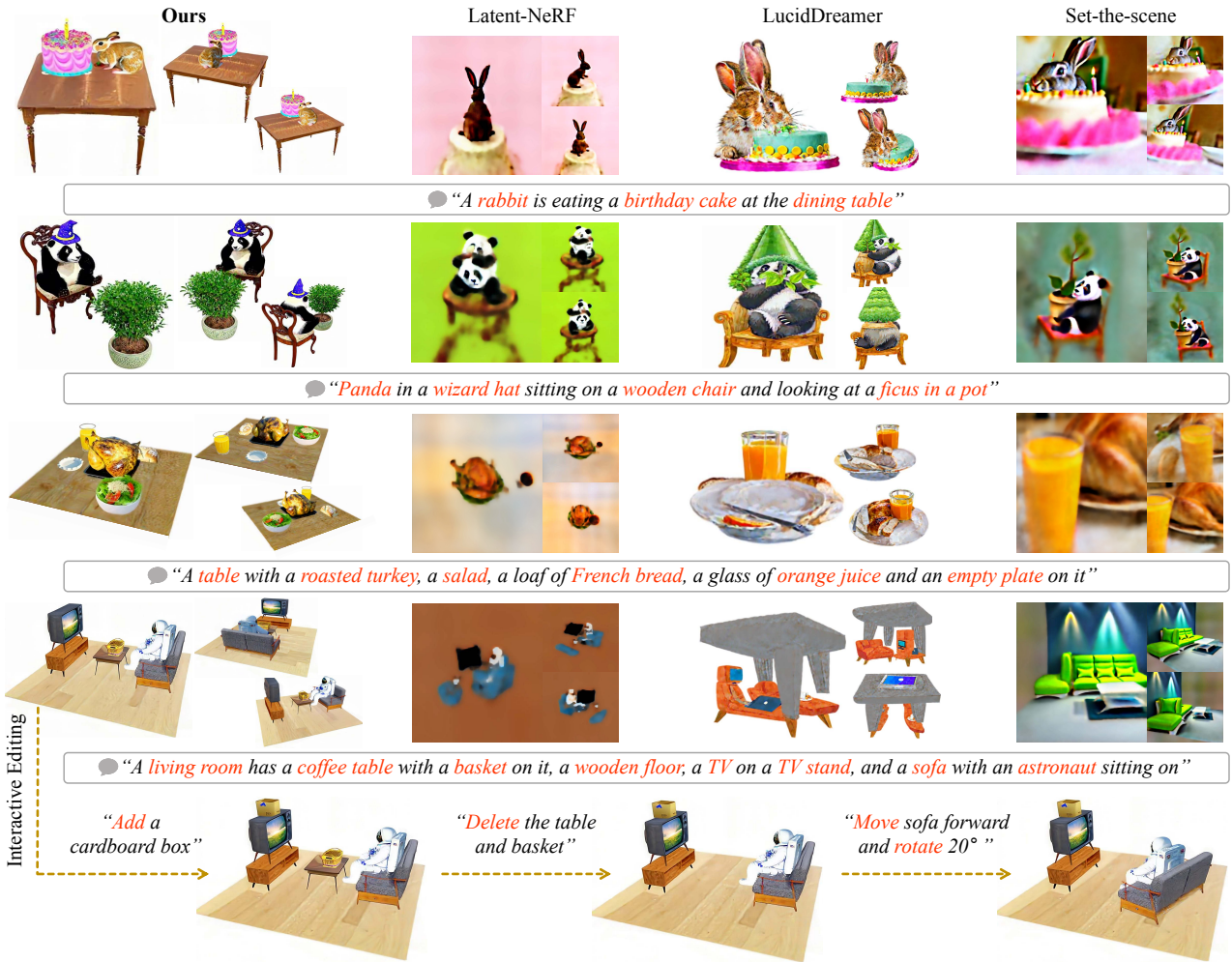}
  \vspace{-2mm}
  \captionof{figure}{\textbf{\ourmethod{}} generates high-quality complex 3D scenes and supports interactive controllable editing. 
  Existing methods either produce low-quality textures, visual artifacts, and geometric distortions or fail to accurately generate multiple objects and their interactions according to the text. 
  }
  \label{fig:teaser}
\end{center}
}
]

\printAffiliationsAndNotice{}

\begin{abstract}
We present \ourmethod{}, generative 3D GAussians with LAyout-guided control, for effective compositional text-to-3D generation. 
We first utilize large language models (LLMs) to generate the initial layout and introduce a layout-guided 3D Gaussian representation for 3D content generation with adaptive geometric constraints.
We then propose an instance-scene compositional optimization mechanism with conditioned diffusion to collaboratively generate realistic 3D scenes with consistent geometry, texture, scale, and accurate interactions among multiple objects while simultaneously adjusting the coarse layout priors extracted from the LLMs to align with the generated scene.
Experiments show that \ourmethod{} is a user-friendly, end-to-end framework for state-of-the-art scene-level 3D content generation and controllable editing while ensuring the high fidelity of object-level entities within the scene.
The source codes and models will be available at \url{gala3d.github.io}.
\end{abstract}

\vspace{-6mm}
\section{Introduction}
\label{Introduction}
Crafting 3D content has been labor-intensive for domain specialists (e.g., 3D artists and interior designers), particularly for complex 3D scenes.
Furthermore, the diversity of the generated scenes remains limited, and ordinary users usually find it challenging to customize scenes or edit them.

These issues have prompted the recent emergence of text-to-3D generation models~\cite{chang2015text, poole2022dreamfusion, lin2023magic3d, raj2023dreambooth3d}.
Given a textual description as input, text-to-3D methods optimize the 3D representations under the supervision of pre-trained 2D diffusion priors, producing object-centric 3D contents~\cite{poole2022dreamfusion, chen2023fantasia3d, xu2023dream3d, wang2023prolificdreamer, tang2023dreamgaussian}. 

However, existing text-to-3D generative models struggle to generate complex 3D scenes with multiple objects and intricate interactions because they optimize a shared 3D representation.
They lack guidance on interactions and spatial positions of objects and generate low-quality 3D scenes, including distorted geometry, 3D inconsistency, multi-face objects, and content drift across different rendering views. 

One recent trend is to introduce manually designed layouts to enforce geometric constraints and capture interactions among multiple objects in the scenes~\cite{po2023compositional, lin2023componerf, cohen2023set}.
However, the implicit NeRF representation~\cite{mildenhall2020nerf} often cannot satisfy all the constraints from the layout, resulting in textural blurring and geometric distortions.
Further, the layout creation requires manual work, which may be time consuming and not friendly for ordinary users.

In this paper, we propose \ourmethod{}, a generative layout-guided Gaussian Splatting framework for complex text-to-3D generation. 
Instead of handcrafted layouts, \ourmethod{} utilizes large language models (LLMs) to extract instance relationships from textual descriptions and translate them into coarse layouts.
We introduce a layout-guided Gaussian representation and adaptively optimize the shape and distribution of Gaussians for high-quality geometry.
Further, we integrate a compositional optimization strategy combined with diffusion priors to update the parameters of layout-guided Gaussians, which enforces semantic and spatial consistency among multiple objects.
To address the misalignment between layouts generated by LLMs and the generated scene, we iteratively optimize the spatial position and scale of the layouts. 

\ourmethod{} presents a user-friendly, end-to-end framework for high-quality scene-level 3D content generation and controllable editing. 
Extensive qualitative and quantitative studies show that \ourmethod{} attains impressive results on compositional text-to-3D scene generation while ensuring high fidelity of object-level entities within the scene.

We make the following contributions in this paper:
\begin{compactitem}
\item We introduce \ourmethod{}, a scene-level text-to-3D framework based on generative 3D Gaussian Splatting, which generates high-fidelity, coherent, complex 3D scenes with multiple objects and precise interactions.
\item \ourmethod{} bridges text description and compositional scene generation through layout priors obtained from LLMs and a layout refinement module that optimizes the coarse layout interpreted by LLMs.
\item \ourmethod{} introduces a layout-guided Gaussian representation with adaptive geometry control to model complex 3D scenes and utilizes a compositional optimization mechanism to tackle the challenge of maintaining 3D consistency in geometry and texture, obtaining accurate interactions among multiple objects.
\item \ourmethod{} outperforms existing methods in text-to-3D scene generation and provides a user-friendly, end-to-end framework for high-quality complex 3D content generation and controllable editing conversationally.
\end{compactitem}

\begin{figure*}[t]
  \centering
  \includegraphics[width=.9\linewidth]{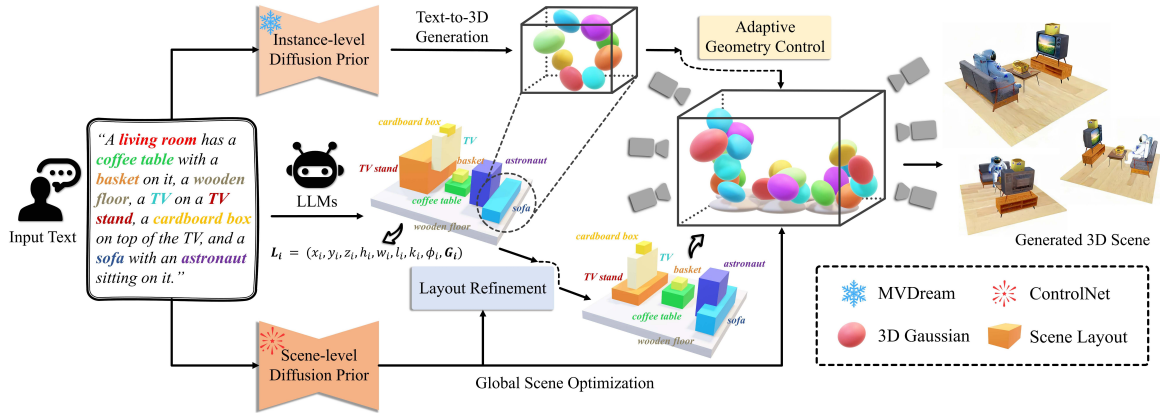}
  \caption{\textbf{Overview of our method.} Given a textual description, \ourmethod{} first creates a coarse layout using LLMs.
The layout is then utilized to construct the Layout-guided Gaussian Representation, incorporating Adaptive Geometry Control to constrain the Gaussians' geometric shape and spatial distribution. 
Subsequently, Compositional Diffusions are employed to optimize the 3D Guassians using text-to-image priors compositionally. 
Simultaneously, the Layout Refinement module refines the initial layout provided by LLMs, enabling better adherence to real-world scene constraints.
}
\label{fig:overview}
\vspace{-4mm}
\end{figure*}

\section{Related Work}
\label{Related Works}

\textbf{Text-to-3D generation by Neural Radiance Field.}
The success of text-to-image methods has been extended to text-to-3D generation, resulting in rapid progress.
DreamFusion~\cite{poole2022dreamfusion} first introduces the Score Distillation Sampling (SDS) to optimize NeRF representations from a pre-trained 2D diffusion model. Magic3D~\cite{lin2023magic3d} improves Dreamfusion with a coarse-to-fine optimization scheme.
In contrast, Fantasia3D~\cite{chen2023fantasia3d} disentangles the modeling of geometry and appearance.
To deal with issues of over-smoothing and out-of-distribution that arise in the diffusion process, ProlificDreamer~\cite{wang2023prolificdreamer} introduces a principled particle-based variational framework named Variational-Score-Distillation, while SJC~\cite{wang2023score} proposes the Perturb-and-Average Scoring.
Recent works~\cite{xu2023dream3d, metzer2023latent} incorporate additional explicit 3D shape priors to assist in generating high-quality 3D geometric structures and assets.
However, the implicit NeRF representation is often insufficient to generate complex scenes that involve multiple objects with intricate interactions. 

One promising approach to address these issues is to use layout to constrain the NeRF representation for compositional 3D generation.
For example, given the user-defined bounding boxes with corresponding texts, Comp3d~\cite{po2023compositional} blends multiple objects into a scene.
Similarly, Set-the-scene~\cite{cohen2023set} and CompoNeRF~\cite{lin2023componerf} generate 3D scenes with compositional NeRFs using pre-defined customizable layouts as object proxies.
However, the layout is manually designed to align with text descriptions, which is time-consuming.
LI3D~\cite{lin2023towards} and SceneWiz3D~\cite{zhang2023scenewiz3d} propose using LLMs as a layout interpreter and connect them to off-the-shelf NeRF-based layout-to-3D generative models~\cite{lin2023componerf} to generate 3D scenes.
However, layouts interpreted by LLMs are often not precise, resulting in misalignment between the layout and the desired scene (e.g., a floating hat, as shown in Figure~\ref{fig:ablationstudy}).
Besides, compositional NeRF models tend to suffer from degradations in visual quality and geometric deformation because they cannot effectively handle the constraints imposed by layout during the NeRF optimization process, as shown in Figure~\ref{fig:teaser}.
Here, we adaptively refine the layout interpreted by LLMs to resolve spatial ambiguities and introduce layout-guided Gaussians to model complex 3D scenes. 

\textbf{Text-to-3D generation by 3D Gaussian Splatting.}
More recently, 3D Gaussian Splatting~\cite{kerbl20233d} (3DGS) provides an efficient point-based representation by optimizing a collection of 3D Gaussian spheres to characterize the 3D space. 
Recent advances have shown promise in merging 3DGS with diffusion models for text-to-3D generation.
\citet{yi2023gaussiandreamer} and \citet{liang2023luciddreamer} utilize 3D text-to-point generative models to generate the initialized point clouds with human priors for 3DGS. 
In contrast, ~\cite{chen2023text, tang2023dreamgaussian} adopt a two-stage optimization process for 3DGS involving geometry optimization and texture refinement.
To maintain multi-view geometric consistency, GaussianDiffusion~\cite{li2023gaussiandiffusion} proposes a variational 3DGS combined with structured noise. 

However, these object-centric methods optimize a single set of 3DGS and cannot effectively generate complex composite scenes with multiple objects.
Further, as there is no constraint on the distribution and shape of Gaussians, these methods may generate distorted geometry, multi-face, and content drift across different rendered views.
To address these issues, we introduce layout priors and adaptive geometry control to make 3DGS more controllable.
Our method expands the capabilities of 3DGS for representing complex multi-object scenes in a compositional construction manner, resulting in high-quality and consistent 3D scene content.

\textbf{3D generation with Large Language Models.}
LLMs possess rich knowledge of large text corpus and can interpret and extract object relationships according to the prompts.
However, this capability has not been extensively explored in the field of 3D generation. 
Some efforts have attempted to leverage LLMs for procedural 3D modeling~\cite{sun20233d}, avatars simulation~\cite{ren2023make}, text-to-3D benchmark~\cite{he2023t}, and 3D editing~\cite{fang2023gaussianeditor}.
The recent combination~\cite{wen2023anyhome, yang2024holodeck, feng2024layoutgpt} of 3D asset retrieval and LLMs has enabled the creation of restricted indoor scenarios. However, these methods have not explored the capability of LLMs in zero-shot 3D generation and end-to-end complex scene structuring.
Furthermore, the aforementioned methods assume that the outputs by LLMs are reliable, leading to potential error propagation with the generated layouts diverging significantly from real-world scenes and textual descriptions. 
\ourmethod{} addresses this by refining LLM-generated layouts to better align with the generated scenes in 3D space, integrating the 3D generation process with layout optimization.

\section{Method}
\label{Method}
As shown in Figure~\ref{fig:overview}, given a text input, \ourmethod{} first obtains coarse layout prior interpreted by LLMs and constructs Layout-guided Gaussian Representation based on the layout (Section~\ref{layout-guided gaussian}). Adaptive Geometry Control is introduced to optimize the geometry and distribution of Gaussian ellipsoids, making them more regularized and closely adherent to the geometric surface (Section~\ref{AGC}). Subsequently, \ourmethod{} utilizes a Compositional Optimization strategy with Diffusion Priors (Section~\ref{CODP}) for layout-guided Gaussians, combined with Layout Refinement module (Section~\ref{DLA}) to refine the coarse layout from LLMs. Our method ultimately employs an aggregated loss function to jointly optimize the entire pipeline (Section~\ref{totalloss}).

\subsection{Layout-guided Gaussian Representation}
\label{layout-guided gaussian}
A few generative models use geometric priors (e.g., layout) to learn 3D representations and ensure shape and consistency.
However, existing methods~\cite{lin2023componerf, po2023compositional, cohen2023set} face two challenges: (i) how to obtain relatively reasonable layout priors without manual design, and (ii) how to mitigate the interference of layout constraints in optimizing 3D representations, minimizing visual artifacts and geometric distortions.

\textbf{Coarse Layout prior interpreted by LLMs.} To deal with the first challenge, we introduce LLMs (e.g., GPT-3.5) as coarse layout interpreters.
LLMs have showcased remarkable language understanding and relationship extraction capabilities, making layout extraction more efficient and cost-effective than manual crafting. 
We utilize LLMs to extract instances from textual descriptions and generate their corresponding coarse layout priors.
Notably, the layout interpreted by LLMs still deviates from the textual descriptions and actual scenes. Therefore, we introduce the Layout Refinement module to address this issue in Section~\ref{DLA}.

\textbf{Layout-guided Gaussian Representation.} For the second challenge, we introduce layout constraints into 3DGS representation for the first time and propose Layout-guided Gaussian Representation.
At a macro level, Layout-guided Gaussian Representation is a collection of scene Gaussians formed by multiple instance Gaussians corresponding to each instance layout. At a micro level, we employ the Adaptive Geometry Control (Section~\ref{AGC}) to better constrain the geometry and distribution of Gaussians. Each set of layout-guided Gaussians can be parameterized as:
\aroundeqn
\begin{equation}
  L_i \!=\! \left \{(x_i,y_i,z_i,h_i,w_i,l_i,k_i,\phi_i, G_i), i \in \left[1,\ldots,N \right] \right \}, 
\end{equation}
\aroundeqn
where $x_i,y_i,z_i$ are the position of layout center for the $i$-th object;
$h_i, w_i, l_i$ represent the length, width, and height of the layout boundary, respectively;
$k_i$ is the scaling factor; $\phi_i$ is the rotation angle; $G_i$ denotes instance Gaussians within the layout; 
and $N$ is the total number of instances in the scene enumerated by LLMs.

Instance Guassians represent the 3D instance through a set of anisotropic Gaussians, defined by center position $\mathbf{p}=(p_x,p_y,p_z) \in \mathbb{R}^{3}$, color $c$, opacity $\alpha$, and covariance $\mathbf{\Sigma_{obj}}=\mathbf{R}\mathbf{S}\mathbf{S}^{\top}\mathbf{R}^{\top}$, 
where $\mathbf{S}$ is the scale matrix and $\mathbf{R}$ is the rotation matrix. 
The scene Gaussians can be then defined as a set of layout-guided Guassians $L_{\mathrm{scene}} = \{ L_i, i \in \left[1,\ldots, N \right] \}$ within the entire scene.

\textbf{Layout-guided Gaussians Rendering at Scene-level.}
To render the entire scene from the Layout-guided Gaussian Representation, we first transform the Gaussians of each layout $L_i$ into a uniform global scene coordinate system:
\aroundeqn
\begin{equation}
\mathbf{p_{\mathrm{scene}}} = k_i \mathbf{R_{z}(\phi_{i})} \mathbf{p_{i}} + (x_{i}, y_{i}, z_{i})^{\top},
\end{equation}
\aroundeqn
where $\mathbf{p}_i$ denotes the center position of instance Gaussians $\mathbf{G_i}$, $k_i$ is the scaling factor, and $\mathbf{R_{z}(\phi_{i})}$ is the rotation matrix for rotating $\phi_i$ degrees around the z-axis. The global scene parameters comprise the covariance matrix of Gaussian collections from all layouts in the scene under transformations:
\aroundeqn
\begin{align}
\mathbf{\Sigma_{\mathrm{scene}}} = k_{i}^{2} \mathbf{R_{z}(\phi_{i})} \mathbf{\Sigma_{obj}} \mathbf{R_{z}^{\top}(\phi_{i})},
\end{align}
\aroundeqn
where $\mathbf{\Sigma_\mathrm{scene}}, \mathbf{\Sigma_\mathrm{obj}}$ represent the covariance of the scene Gaussians and instance Gaussians; $\mathbf{R_{z}}$ is the rotation matrix and $\mathbf{R_{z}}^{\top}$ is its transpose. 
The corresponding 2D covariance can be projected by:
\aroundeqn
\begin{equation}
\mathbf{\Sigma_{\mathrm{scene}}^{\prime}} = \mathbf{J} \mathbf{W} \mathbf{\Sigma_\mathrm{scene}} \mathbf{W^{\top}} \mathbf{J^{\top}},
\end{equation}
\aroundeqn
where $\mathbf{W}$ is the viewing transformation matrix and $\mathbf{J}$ denotes the Jacobian of the affine approximation of the projective transformation.
We further utilize global Gaussian splatting to render the entire scene containing multiple objects:
\aroundeqn
\begin{equation}
  C = \sum_{i\in N} c_i \alpha_{i}^{\prime} \prod_{j=1}^{i-1} (1-\alpha_{j}^{\prime}),
\end{equation}
\aroundeqn
where $C$ is the color of the rendering pixel;
$c_i$ is the rendering color of each Gaussian; and $\alpha_{i}^{\prime}$ is the final opacity of the Gaussian. 
The final opacity $\alpha_{i}^{\prime}$ is queried by $\mathbf{Q}$, the rendering pixel's coordinate in the projection space:
\begin{equation}
  \alpha_{i}^{\prime} = \alpha_{i} e^{-\frac{1}{2}(\mathbf{Q}-\mathbf{P_i})^{\top}\mathbf{\Sigma_{i}^{-1}}(\mathbf{Q}-\mathbf{P_i)}},
\end{equation}
where $\mathbf{\Sigma_{i}^{-1}}$ is equivalent to the axes of the ellipsoid; $\alpha_{i}$ denotes the learned opacity; and $\mathbf{P_i}$ is the spatial position of the Gaussian in the projected plane.

\subsection{Adaptive Geometry Control for Gaussians}
\label{AGC}
The raw 3DGS representation adopts the densification scheme for Gaussians, providing good control over the total number of Gaussians. However, this strategy fails to constrain the distribution of Gaussian ellipsoids, resulting in numerous unused invisible Gaussians. It is also incapable of controlling the generation of Gaussians with a uniform regular shape, which shares similar covariances and normal vectors. As a comparison, we propose Adaptive Geometry Control for Gaussians, which achieves adaptive geometric control of Layout-guided Gaussians through distribution constraint and shape optimization.
Similar to \cite{compression3dpoint, liu2023humangaussian, low2023robust}, given an initialized set of Gaussians, the distribution constraint can be implemented by a density distribution function:
\aroundeqn
\begin{equation}
  \frac{1}{\Vert \mathbf{p_i}- \mathbf{\zeta_i} \Vert} \sim \mathcal{\hat{N}} (\mu, \sigma^2),
  \label{eq:density}
\end{equation}
\aroundeqn
where $\mathbf{\zeta_i} = (x_i, y_i, z_i)$ is the center coordinate of the corresponding layout prior; $\Vert \mathbf{p_i}- \mathbf{\zeta_i} \Vert$ is the Euclidean distance between the Gaussian center and $\mathbf{\zeta_i}$; $\mu$ is the mean of the Gaussians' distributions; and $\sigma$ is the standard deviation. 
Both are changeable parameters.
Here, $\mathcal{\hat{N}}$ represents the folded normal distribution, with a truncation range from the layout center to the boundary.
We then sample Gaussians near the layouts' surface according to the distribution.

\begin{figure}[h]
  \centering
  \includegraphics[width=\linewidth]{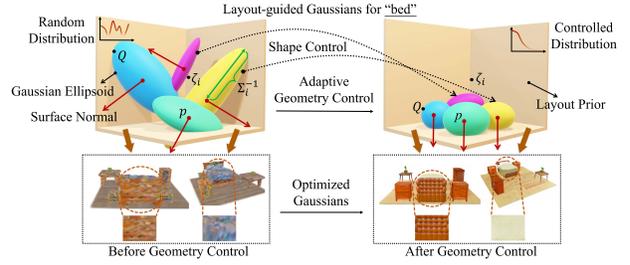}
  \caption{\textbf{Adaptive Geometry Control for instance Gaussians.} Note that the improved Gaussian distribution results in enhanced texture and geometry, as the colors of Gaussians on the surface become more aligned.
  }
  \label{fig:control}
  \vspace{-10pt}
\end{figure}

To obtain Gaussian shapes with more regular geometry and scale, we introduce a regularization term:
\aroundeqn
\begin{equation}
  \mathcal{L}_{reg} = \frac{1}{N}\sum_{i=1}^{N} \mathbf{S_i} \left\Vert \mathbf{q}-\mathbf{p_i} \right\Vert,
\end{equation}
\aroundeqn
where $\mathbf{S_i}$ is a 3D vector along three axes and denotes the scale matrix for the $i$-th Gaussian; and $\mathbf{q}-\mathbf{p_i}$ denotes the flatness of the Gaussian ellipsoid and will be compressed if too long.
As shown in Figure~\ref{fig:control}, Adaptive Geometry Control adaptively optimizes the distribution and shape of the layout-guided Gaussians, achieving more refined geometric structures and highly detailed textures.

\subsection{Compositional Optimization with Diffusion Priors}
\label{CODP}
In pursuit of generating scenes with a consistent style and multiple instances, our method leverages a compositional optimization strategy with diffusion priors to update the parameters of Layout-guided Gaussians.
We initially utilize a multi-view diffusion model to optimize instance Gaussian, followed by a scene-conditioned diffusion to align and optimize multiple objects in the scene along with their interactive relationships. Layout loss is further employed to ensure the semantic and spatial consistency between the generated 3D scene and the layout prior.

\textbf{Text-to-3D generation by multi-view diffusion.}
To optimize instance Gaussian for each instance in the scene, we utilize MVDream~\cite{shi2023mvdream} as a multi-view diffusion prior combined with Score Distillation Sampling (SDS). The gradient for the $i$-th instance Gaussian can be formulated as:
\aroundeqn
\begin{equation}
  \bigtriangledown_{G_{i}} \mathcal{L}_{SDS} ^{(i)}=\mathbb{E}_{\epsilon ,\eta} \left[ w(\eta) (\epsilon_{\varphi}(I_{i};t_{i},M_{i},\eta)-\epsilon )\frac{\partial I_{i}}{\partial G_{i}} \right],
\end{equation}
\aroundeqn
where $\epsilon$ is the added noise;
$t_i$ is the text prompt corresponding to the $i$-th instance; $\eta$ is the time step for optimization; $w(\eta)$ is a weighting function from DDPM~\cite{ho2020denoising}; $I_{i}$ denotes the sampled image from diffusion prior; $M_{i}$ is the extrinsic matrix of the camera; $G_{i}$ denotes the instance Gaussians within the layout, and $\epsilon_{\varphi}$ is the denoising function for the diffusion process of 3DGS.
We embed a virtual camera model to render multi-view images from diffusion prior, with a camera radius of $\frac{3}{4} \Vert (h_i,w_i, l_i) \Vert_{2}$, a horizontal angle of $360^{\circ}$, and uniform sampling of viewing poses.

\textbf{Global scene optimization by conditioned diffusion.} We then introduce conditioned diffusion to optimize the global scene, generating interactions between multiple instances while adhering to the layout prior. Unlike single object generation, we use ControlNet~\cite{zhang2023adding} for compositional optimization, ensuring that the generated scene aligns with the layout. Concretely, we fine-tuned ControlNet to support rendering layouts from multiple viewpoints as input and generate 2D diffusion supervision with layout-text consistency. The gradients of SDS for scene parameters can be formulated as:
\aroundeqn
\begin{equation}
\resizebox{0.88\hsize}{!}{
  $ \bigtriangledown_{G_{\mathrm{scene}}} \mathcal{L}_{SDS} = \mathbb{E} _{\epsilon ,\eta} \left[  w(\eta) (\epsilon_{\phi}(I;t,\delta,\eta)-\epsilon )\frac{\partial I}{\partial G_{\mathrm{scene}}} \right]$,
  }
\end{equation}
\aroundeqn
where $\delta$ is the condition input for the ControlNet, obtained by rendering the 2D images from the layouts. During the diffusion process, the instance-level and scene-level optimization share the same time step $\eta$ to ensure synchronous and collaborative learning. $t$ is the textual description of the whole scene encompassing multiple instances; $I$ is the rendered global scene from conditioned diffusion, and $G_{scene}$ denotes the parameters of scene Gaussians. 

\textbf{Global scene optimization by Layout loss.} To constrain the generated instances in 3D space to maintain scale, position, and geometric consistency with the provided layout priors, we introduce the layout loss:
\aroundeqn
\begin{equation}
\resizebox{0.88\hsize}{!}{
   $\mathcal{L}_{layout}^{(i)} = \vmathbb{1}_{bbox}(\mathbf{p}) [ d^x (p_{x},x_{i},h_{i}) + d^y (p_{y},y_{i},w_{i}) +  d^z (p_{z},z_{i},l_{i}) ]$,
   }
\end{equation}
\aroundeqn
where distance function $d^x (p_{x},x_{i},h_{i})$ calculates the Manhattan distance from each center point outside the 3D layout boundaries to the nearest point on the x-axis and similarly for the other two axes:
\aroundeqn
\begin{equation}
\resizebox{0.88\hsize}{!}{
 \!\!\!\!\!  $ d^{x} (p_{x}, x_{i}, h_{i})  \!=\! \min( | p_{x} - (x_{i} + \frac{h_{i}}{2} )  | , | p_{x} - ( x_{i} - \frac{h_{i}}{2} ) | )$,
}
\end{equation}
\aroundeqn
where $p_x$ is the center coordinate of the instance Gaussian on the x-axis, $x_i$ is the position of the layout center on the x-axis, and $h$ is the height of the layout prior to the $i$-th instance. The indicator function $\vmathbb{1}_{bbox}(\mathbf{p})$ checks whether a point $\mathbf{p}$ is in the bounding box and is $1$ if $\ p_x \in [x_{i} - \frac{h_{i}}{2},  x_{i} + \frac{h_{i}}{2}]$ and $\ p_y \in [y_{i} - \frac{w_{i}}{2},  y_{i} + \frac{w_{i}}{2}]$ and $\ p_z \in [z_{i} - \frac{l_{i}}{2},  z_{i} + \frac{l_{i}}{2}]$, and $0$ otherwise.
\begin{figure}[h]
  \centering
\includegraphics[width=\linewidth]{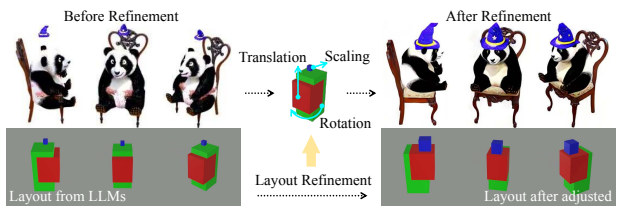}
  \caption{\textbf{Layout Refinement.} The LLM-generated layouts exhibit spatial misalignment and abnormal scale. We employ Layout Refinement to optimize the layout, resulting in a more aligned layout with the text and the 3D scene.}
  \label{fig:refine}
  \vspace{-10pt}
\end{figure}

\subsection{Layout Refinement}
\label{DLA}
Although LLMs possess the ability to extract textual-instance relationships, they may still exhibit significant errors due to the lack of 3D understanding of scenes. The layout priors interpreted by LLMs may deviate from the actual scene and text description, leading to issues like object drift and size discrepancies (Figure~\ref{fig:refine}). To solve this issue, we propose the Layout Refinement module to adaptively adjust the coarse layout generated from LLMs, making it more consistent with scene constraints. 
The gradient of the layout can be formulated as:
\aroundeqn
\begin{equation}
\resizebox{0.88\hsize}{!}{
  $\bigtriangledown_{(\zeta_i,\alpha_{i},k_{i},\phi_i)}  \mathcal{L}_{Def}^{(i)} =
  \mathbb{E}_{\epsilon,\eta} [ w(\eta) (\epsilon_{\phi}(I;t,\delta,\eta)-\epsilon )\frac{\partial I}{\partial (\zeta_i,\alpha_{i},k_{i},\phi_i)} ]$,
  }
\end{equation}
\aroundeqn
where $\zeta_i = (x_i,y_i,z_i)$ is the center coordinate of the layout corresponding to the $i$-th instance; $\alpha_{i}$ is the opacity; $k_{i}$ denotes the scale scaling factor, and $\phi_{i}$ is the rotation matrix for the layout. All of the above are learnable parameters, continuously updated during the optimization. $t$ is the text prompt, and $I$ denotes the rendered scene image from the conditioned diffusion priors.

\subsection{Total Loss}
\label{totalloss}
The total loss function can be summarized as
\begin{equation}
\resizebox{0.88\hsize}{!}{
  $ \mathcal{L} =\sum_{i=1}^{N} ( \beta_{1} \mathcal{L}_{SDS} ^{(i)} + \beta_{2} \mathcal{L}_{layout} ^{(i)}  + \beta_{3} \mathcal{L}_{Def}^{(i)} ) + \beta_{4} \mathcal{L}_{global}  + \beta_{5} \mathcal{L}_{reg},$
  }
\end{equation}
where $\mathcal{L}_{SDS}^{(i)}$ optimize the $i$-th layout-guided instance Gaussian, $\mathcal{L}_{layout}^{(i)}$ optimize the corresponding $i$-th layout. $\mathcal{L}_{Def}^{(i)}$ denotes the Layout Refinement for coarse layout priors. $\mathcal{L}_{global}$ denotes the global optimization by conditioned diffusion for the whole scene, and $\mathcal{L}_{reg}$ is applied to supervise the shape control for Gaussians.

\section{Experimental Results}
\label{Experiment}

\begin{table*}[!t]
  \caption{
  {\textbf{Overall performance of \ourmethod{} with existing state-of-the-art Text-to-3D approaches using single-object and multi-object text prompts.} \textbf{T} denotes using text prompt and \textbf{TL} denotes using text prompt combined with layout. \case, \casee, \caseee, \caseeee \ refer to the number of instances in the scene as 1, 3, 7, and 10, respectively. Average represents the average score of multiple generated scenes used for evaluation, including 22 scenes with varying numbers of objects, ranging from 1 to 10.}}
  \centering
  \scriptsize
  \setlength{\tabcolsep}{3.5mm}{
    \begin{tabular}{c|ccccccc}
    \textbf{Methods} & \textbf{Representation} & \textbf{Input} & \textbf{Average} & \textbf{\case Case 1} & \textbf{\casee Case 2} & \textbf{\caseee Case 3} & \textbf{\caseeee Case 4}\\
    \hline
    Latent-NeRF~\cite{metzer2023latent}                & NeRF                           & T  & 27.772 & 22.135 & 27.482 & 22.203 & 19.606 \\
    ProlificDreamer~\cite{wang2023prolificdreamer}     & NeRF                           & T  & 28.401 & \cellcolor{tabsecond} 30.237 & 21.913 & 19.219 & 25.587 \\
    MVDream~\cite{shi2023mvdream}                      & NeRF                           & T  & \cellcolor{tabthird} 30.856 & 28.756 & \cellcolor{tabsecond} 32.636 & 26.015 & \cellcolor{tabsecond} 27.417 \\
    SJC~\cite{wang2023score}                           & Voxel Grid                     & T  & 28.775 &  29.100 & \cellcolor{tabthird} 31.764 & 21.154 & 26.352 \\
    \hline
    DreamGaussian~\cite{tang2023dreamgaussian}         & 3DGS                           & T  & 25.117 & 26.281 & 23.051 & 18.595 & 25.739 \\
    GaussianDreamer~\cite{yi2023gaussiandreamer}       & 3DGS                           & T  & 28.351 & \cellcolor{tabthird} 29.469 & 31.237 & 25.727 & 24.143 \\
    GSGEN~\cite{chen2023text}                          & 3DGS                           & T  & 30.293 & 28.932 & 29.578 & \cellcolor{tabsecond} 29.959 & 23.927 \\
    LucidDreamer~\cite{liang2023luciddreamer}          & 3DGS                           & T  & \cellcolor{tabsecond} 31.174 & 28.720 & 26.533 &  27.768 & \cellcolor{tabthird} 26.895 \\
    \hline
    Set-the-scene~\cite{cohen2023set}                  & Comp NeRF             & TL     &  29.628  &  28.129   & 19.135 & \cellcolor{tabthird} 29.003 & 25.899 \\
    \hline
    \textbf{Ours}                                     & Comp 3DGS             & T  & \cellcolor{tabfirst} 34.573 & \cellcolor{tabfirst} 31.637  & \cellcolor{tabfirst} 37.658 & \cellcolor{tabfirst} 31.459 &  \cellcolor{tabfirst} 35.052 \\
    \end{tabular}
    }
  \label{compare_SOTA}
  \aftertab
\end{table*}

\paragraph{Implementation details.}
We utilize MVDream~\cite{shi2023mvdream} as the multi-view diffusion model, with a guidance scale of 50. The guidance scale of ControlNet is set to 100 to optimize the scene and decrease the timestep linearly during training. For the 3DGS, the learning rates of opacity and position are $5 \times 10^{-2}$ and $1.6 \times 10^{-4}$. The color of 3D Gaussians is represented by the spherical harmonic coefficient, with the degree set to 0 and the learning rate set to $5 \times 10^{-3}$. The covariance of the 3D Gaussians is converted into scaling and rotation for optimization, with learning rates of $5 \times 10^{-3}$ and $10^{-3}$, respectively. We set coefficients ${\beta_1, \beta_2, \beta_3, \beta_4}$ as $\beta_1 = 1, \beta_2 = 10^3, \beta_3 = 10^{-1}, \beta_4 = 10^{-1}$, and $\beta_5 = 10^{3}$ to balance the magnitude of the losses.
For each instance, we initialize the 3D Gaussians with 100,000 particles and discard adaptive density control in 3D Gaussian Splatting to save memory and speed up training. The sampling radius of the camera is set to the scene range in the spherical coordinate system, while horizontal angles are uniformly sampled at $360^{\circ}$. All the experiments are carried out on a single A800 with 80 GB memory.

\subsection{Quantitative Comparison}
To evaluate our method on the Text-to-3D task, we conduct benchmarking against the state-of-the-art (SOTA) approaches, including NeRF-based methods~\cite{metzer2023latent, wang2023prolificdreamer}, Voxel-based method~\cite{wang2023score}, 3DGS-based methods~\cite{tang2023dreamgaussian, yi2023gaussiandreamer, chen2023text, liang2023luciddreamer}, and compositional NeRF-based generation with layout~\cite{cohen2023set}. Given the absence of ground truth for zero-shot text-to-3D generation, we follow previous works~\cite{jain2022zero, huang2023dreamtime} to employ CLIP Score as the evaluation metric to assess the quality and consistency of generated 3D scenes in relation to textual descriptions.
As shown in Table~\ref{compare_SOTA}, text prompts containing varying numbers of objects are chosen to assess the performance of text-to-3D generative models under different settings.
Our method excels over all competitors in generating complex 3D scenes with multiple interacting objects.

\textbf{Compared with NeRF-based and voxel-based methods.} To ensure a fair comparison, we employ the vanilla form of Latent-NeRF~\cite{metzer2023latent}, which employs a text-guided NeRF model to optimize the spatial radiance field in latent space. Our method outperforms Latent-NeRF by a large margin across all evaluated metrics.
ProlificDreamer~\cite{wang2023prolificdreamer} presents Variational Score Distillation for 3D scene generation, maintaining a set of parameters as particles to represent the 3D distribution. However, it fails to model complex scenes with multiple interacting objects using this scheme.
\ourmethod{} also boosts the performance of our baseline method MVDream~\cite{shi2023mvdream} in both object-level and scene-level generation and achieves optimal results.
SJC~\cite{wang2023score} regards the 3D diffusion process as an optimization of a vector field. Instead, our method integrates conditioned diffusion with compositional optimization of Gaussians, proven to be more effective for 3D scene generation.

\textbf{Compared with compositional NeRFs with scene layout.} Recent works~\cite{cohen2023set, lin2023componerf, po2023compositional} utilize manually designed layouts as priors for compositional NeRF to assist in generating more controllable and intricate scenes. We compare our method with Set-the-scene~\cite{cohen2023set} and provide the rendered layout interpreted by LLMs as its prior input. Under the same specified layout input, scenes generated by Set-the-Scene exhibit unpleasant blurriness and artifacts.
Conversely, our method demonstrates superior scene consistency, spatial geometry, and overall quality, especially in scenes with multiple instances (e.g., ten objects).

\textbf{Compared with 3DGS-based methods.} For Gaussian-based approaches~\cite{tang2023dreamgaussian, yi2023gaussiandreamer, chen2023text, liang2023luciddreamer}, our method exhibits superior performance in 3D generation for both single-object and complex scenes. Our proposed Adaptive Geometry Control for 3DGS ensures the generation of 3D models with high-resolution geometry and texture, thereby avoiding the distortions and blurring observed in existing approaches.

\begin{table*}[!t]
  \centering
  \caption{
  {\textbf{User study results.} Human evaluation results comparing \ourmethod{} with other SOTA text-to-3D approaches. Participants scored on the following four metrics, rating from 1 to 10, with higher scores indicating stronger preference.}}
    \footnotesize 
  \centering
  \setlength{\tabcolsep}{1.5mm}{
  {
    \begin{tabular}{c|cccccc}
    \textbf{Methods} & \textbf{Scene Quality} & \textbf{Geometric Fidelity} & \textbf{Text Alignment} & \textbf{Scene Consistency}\\
    \hline
    SJC~\cite{wang2023score}                           &  5.98   &  5.04 & \cellcolor{tabthird} 6.76 & 4.61  \\
    DreamGaussian~\cite{tang2023dreamgaussian}         &   5.22  & 4.18  &  4.30 &  5.46 \\
    GaussianDreamer~\cite{yi2023gaussiandreamer}       &  6.09   & \cellcolor{tabsecond} 5.71 & 5.23 &  4.37  \\
    GSGEN~\cite{chen2023text}                          &   \cellcolor{tabsecond} 6.54  &  4.23 & 5.41 &  \cellcolor{tabsecond} 6.25  \\
    LucidDreamer~\cite{liang2023luciddreamer}          &   4.78  & \cellcolor{tabthird} 5.62  & 5.03 &  4.77  \\
    Set-the-scene~\cite{cohen2023set}                  &  \cellcolor{tabthird} 6.36  &    5.03   &  \cellcolor{tabsecond} 7.12  & \cellcolor{tabthird} 6.12  \\
    \hline
    \textbf{Ours}                                     &   \cellcolor{tabfirst} 8.42   & \cellcolor{tabfirst} 8.37  & \cellcolor{tabfirst} 8.55 & \cellcolor{tabfirst} 9.68 \\
    \end{tabular}
    }
    }
  \label{user_study}
  \aftertab
\end{table*}

\begin{figure*}[t]
  \centering
  \includegraphics[width=0.9\linewidth]{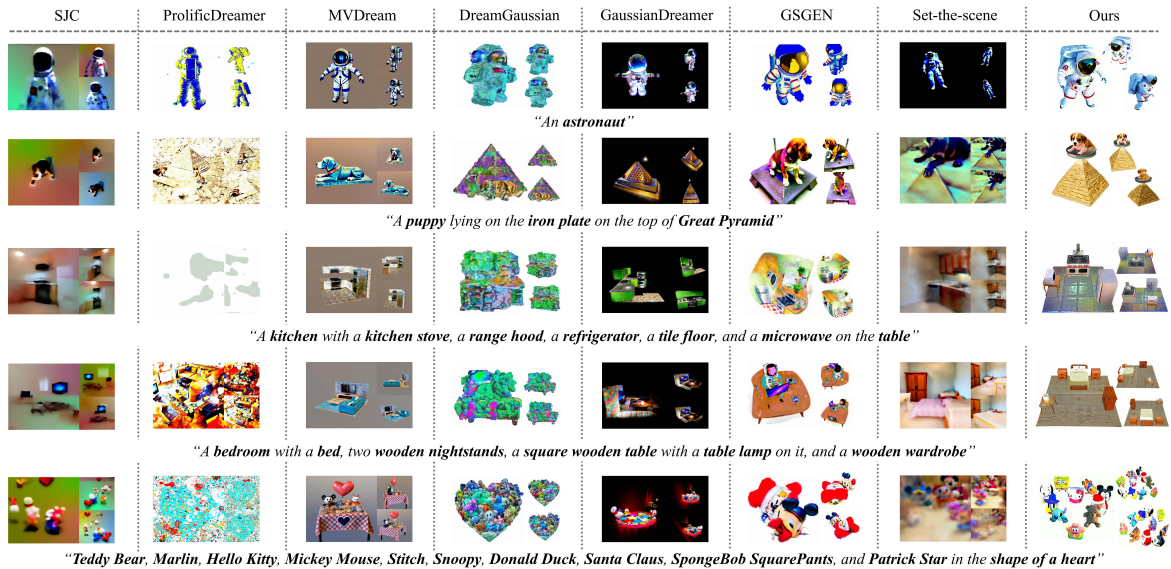}
  \caption{\textbf{Qualitative comparisons of text-to-3D generation approaches.} Our method is capable of generating high-quality single-object, interactive multi-object, and complex composite scenes with high consistency in textual descriptions.
}
  \label{fig:compbig}
\vspace{-2mm}
\end{figure*}

\subsection{Qualitative Comparison}
We report qualitative comparisons on text-to-3D generation in Figure~\ref{fig:teaser} and Figure~\ref{fig:compbig}, including the generation of single-object, interactive multi-object, and complex composite scenes.
Visually, our method enables the generation of highly realistic 3D objects and multi-object scenes, surpassing other methods in terms of generated texture, geometric shapes, and semantic consistency. Notably, the NeRF-based generative models produce noticeable artifacts, distortions, and multi-view inconsistencies. The 3DGS-based methods often exhibit multi-face issues and rough geometric shapes. Additionally, these methods show significant deficiencies in scene-text alignment, struggling to accurately generate specified instances, interaction relationships, and correct spatial positions.
\ourmethod{} not only precisely generates the desired multiple objects and their interaction relationships but also maintains the consistency between text and multiple objects in the scene, ensuring a unified style.

\begin{figure}[!t]
  \centering
  \includegraphics[width=0.9\linewidth]{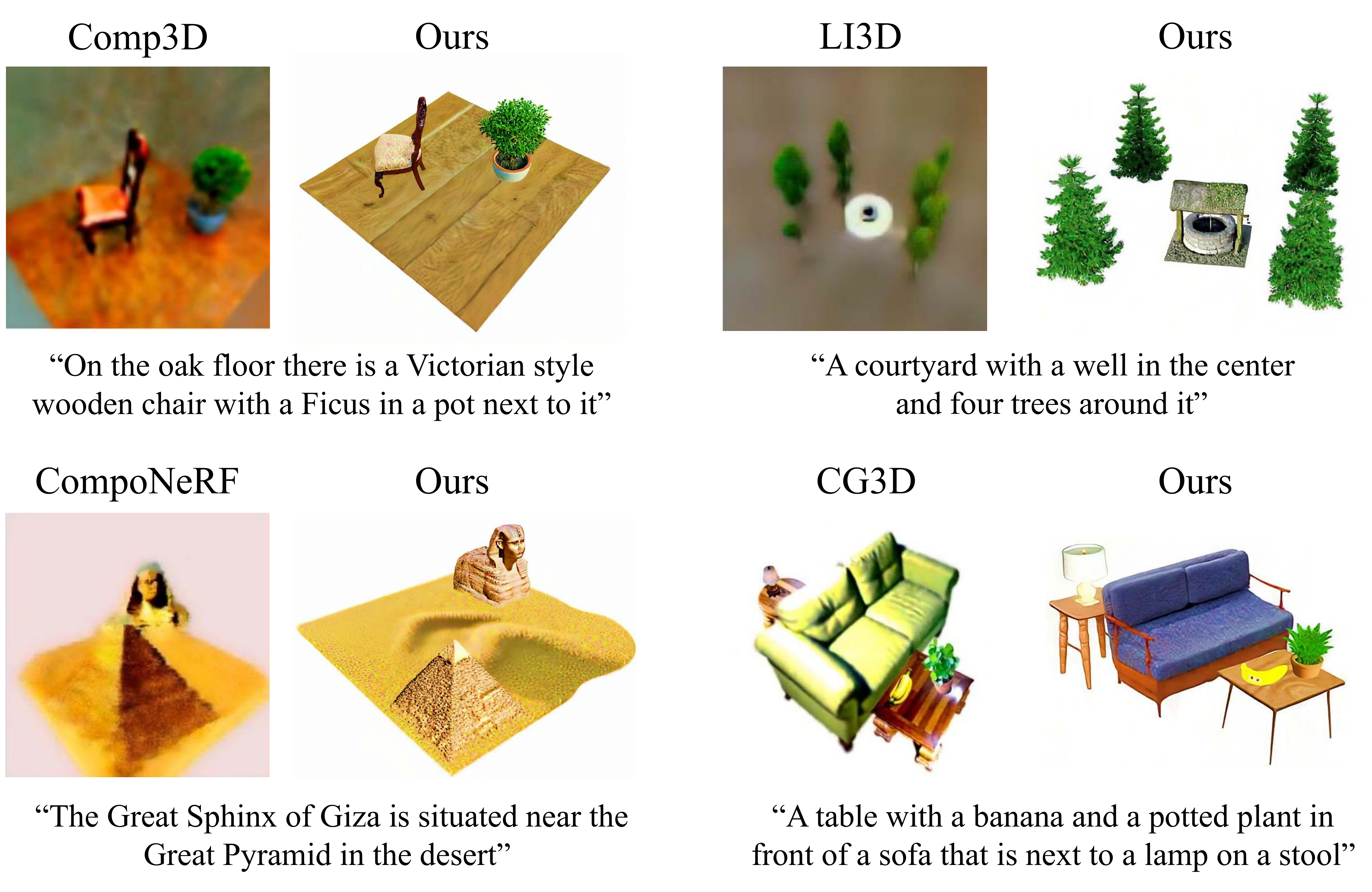}
  \caption{\textbf{Comparisons with compositional scene generation methods.} Our method ensures superior coherence and consistency in generated content compared to competitors.}
  \label{fig:comp2}
\vspace{-5mm}
\end{figure}

\textbf{Compared with compositional scene generation methods.} We further compare our approach with recent works~\cite{lin2023componerf, vilesov2023cg3d, po2023compositional, lin2023towards} in compositional scene generation, which use the layout as an additional constraint for 3D representation (e.g., NeRF).
Since most of these works are not open-sourced, we use the results provided in their papers for comparison, applying the same prompts as input to generate 3D scenes.
As shown in Figure~\ref{fig:comp2}, these compositional 3D scene generation approaches exhibit unpleasant floating objects, visual artifacts, and geometric distortions in the generated 3D scenes. They also face significant challenges in texture degradation and transition smoothness. In contrast, the 3D scenes generated by our method obtain higher realistic visual effects.
Our method also supports more user-friendly and controllable editing in a convenient interactive manner.

\subsection{User Study}
We conduct a user study to further evaluate the effectiveness of our method in generating high-quality, text-consistent 3D assets.
Specifically, we engage human evaluators to compare 3D models generated by our method and competitive approaches from 8 text descriptions.
A total of 125 participants were asked to rank based on four dimensions: (a) Scene Quality, (b) Geometric Fidelity, (c) Text Alignment, and (d) Scene Consistency. Each round of comparison requires participants to rate the four assessment options on a scale from 1 to 10 (10 being the best and vice versa).
Among these users, $39.2\%$ are professionals in the fields of art design and 3D modeling.

We report the average score of the trial, reflecting user preferences for generated 3D assets. As shown in Table~\ref{user_study}, the results demonstrate a clear preference for our method, receiving consistently positive reviews. Compared with SOTA approaches, \ourmethod{} excels across all four assessments.
Our approach also garners preferences from domain experts, demonstrating its potential in practical applications.

\begin{figure}[h]
  \centering
  \includegraphics[width=\linewidth]{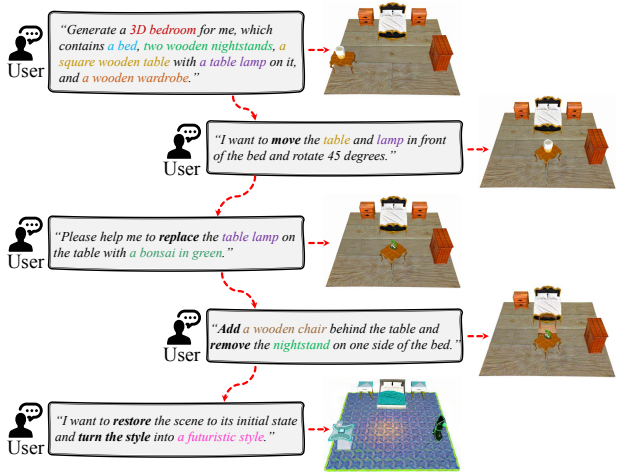}
  \caption{\textbf{Conversational Interactive Editing.} Our method facilitates user-friendly and controlled editing of 3D scenes.}
  \label{fig:chat}
  \vspace{-10pt}
\end{figure}

\subsection{Conversational Interactive Editing}
Our method allows conversational interactive editing. Users can freely and controllably edit the generated scene based on textual conversations. Specifically, editing instructions are initially interpreted by LLMs into corresponding layout transformation operations (e.g., adding/removing objects, moving positions, rotating angles, etc.). We then optimize the Layout-guided Gaussian Representation in the edited local layout areas while maintaining the stability of other regions.
Our approach guarantees highly controllable and personalized scene editing, including the addition or removal of objects, spatial adjustments, style transfer, and object interactions. This paradigm of conversational interactive editing combined with LLMs achieves real-world applications, providing a user-friendly 3D assets generation and customized editing pipeline.

\begin{figure}[h]
  \centering
  \includegraphics[width=\linewidth]{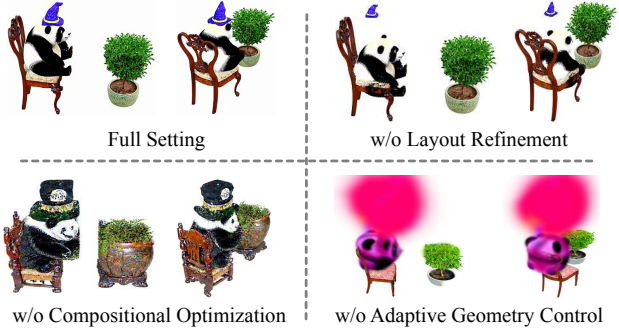}
  \caption{\textbf{Visual results of the ablation studies.} Experiments validate the effectiveness of each proposed module, highlighting the crucial role of Layout-guided Gaussian representation coupled with Adaptive Geometry Control in producing high-quality scene geometry and texture.}
  \label{fig:ablationstudy}
  \vspace{-10pt}
\end{figure}

\subsection{Ablation Studies}
\textbf{Adaptive Geometry Control for Gaussians.} We replace the Adaptive Geometry Control with the density control scheme employed by the raw 3DGS~\cite{kerbl20233d} and observe a significant decrease in the realism of the generated scene, as shown in Figure~\ref{fig:ablationstudy}. The Gaussian densification fails to constrain the distribution and shape of Gaussian ellipsoids, resulting in unpleasant artifacts and blurs. In contrast, our method continuously optimizes the geometric shapes and spatial distributions of 3D Gaussians during the training process. The ablation confirms the effectiveness of Adaptive Geometry Control, which finely improves the complex topological structures and results in enhanced texture and geometry within the global optimization space.

\textbf{Layout Refinement Module for LLMs interpreted coarse layout.} Directly using the layout interpreted by LLM without refinement results in 3D scenes not well aligned, as shown in Figure~\ref{fig:refine} and Figure~\ref{fig:ablationstudy}. By contrast, the Layout Refinement module enables the optimizing of layouts, continuously adjusting them throughout the denoising process to achieve more intricately aligned interactions among instances, adhering closely to real-world constraints.

\begin{table}[!t]
  \centering
  \caption{
  {\textbf{Effect of each module in our proposed method.} AGC is short for Adaptive Geometry Control, LRM denotes the Layout Refinement Module, and COS denotes the Compositional Optimization Scheme.}}
    \footnotesize
  \centering
  \setlength{\tabcolsep}{1.6mm}{
  {
    \begin{tabular}{lclc}
    \textbf{Model} & \textbf{CLIP Score} & \textbf{Model} & \textbf{CLIP Score} \\
    \hline
    w/o AGC                      &  32.198      & w/o LRM                      &  \cellcolor{tabthird} 34.293 \\
    w/o COS                      & 32.213      & w/o $\mathcal{L}_{global}$              & \cellcolor{tabsecond} 34.342          \\    
    w/o $\mathcal{L}_{layout}$             & 33.297    
    &    Ours-Full                    & \cellcolor{tabfirst} 34.885    \\
    \end{tabular}
    }
    }
  \label{Ablation_Study}
  \vspace{-5pt}
  \aftertab
\end{table}

\textbf{Compositional Optimization Scheme.} Figure~\ref{fig:ablationstudy} shows ablations to assess the efficacy of the proposed compositional optimization scheme. 
Specifically, we remove the Global Scene Optimization module, retaining only SDS supervision for instances (with MVDream), and concatenate each object into the scene according to the adjusted layouts. Due to the absence of comprehensive global scene optimization, the generated 3D scenes exhibit impoverished textures and lack scene coherence. Furthermore, the generated geometry only adheres to local layout supervisions, resulting in the emergence of ``over-constrained'' boundaries.

\textbf{Effect of Loss Functions.} We analyze how each proposed loss function contributes to the final performance. As shown in Table~\ref{Ablation_Study}, results indicate that both $L_{layout}$ and $L_{scene}$ improve the generating quality, enhancing texture details and maintaining text-3D alignment.

\section{Conclusion}
\label{Conclusion}

In this paper, we present \ourmethod{}, a scene-level text-to-3D framework based on generative layout-guided 3D Gaussian representation, which generates high-fidelity, 3D consistent scenes with multiple objects. 
Experiments demonstrate that our method surpasses existing methods in text-to-3D generation, showcasing the ability to generate complex scenes with multiple objects and interactions, achieving outstanding texture and geometry.
Our method also facilitates interactive and controllable scene editing, achieving an efficient and user-friendly 3D scene generation and editing framework.

\clearpage

\section*{Acknowledgment}
This work was supported in part by the National Natural Science Foundation of China under Grant 62176007 and China National Petroleum Corporation-Peking University Strategic Cooperation Project of Fundamental Research. This work was also a research achievement of Key Laboratory of Science, Technology, and Standard in Press Industry (Key Laboratory of Intelligent Press Media Technology).

\section*{Impact Statement}
This paper presents work aimed at advancing the fields of Deep Learning and 3D Vision.
While AI-generated 3D content offers numerous advantages, it also introduces adverse social impacts. The automation of 3D modeling and scene creation could pose potential risks to the labor market. Additionally, similar to other generative models, our approach has the capacity to generate deceptive and malicious 3D content, highlighting the importance of exercising caution in its application.

\bibliography{icml}
\bibliographystyle{icml2024}

\end{document}